\documentclass[10pt, twocolumn, letterpaper]{article}

\usepackage{iccv}
\usepackage{times}
\usepackage{epsfig}
\usepackage{graphicx}
\usepackage{amsmath}
\usepackage{subfigure}
\usepackage{amssymb}

\usepackage[pagebackref=true, breaklinks=true, letterpaper=true, colorlinks, bookmarks=false]{hyperref}

\iccvfinalcopy 

\def\iccvPaperID{6} 

\ificcvfinal\fi

\begin{document}

\title{AirFace:Lightweight and Efficient Model for Face Recognition }

\author{Xianyang Li\\
AIRIA\\
China\\
{\tt\small qq1052308104@gmail.com}
\and
Feng Wang\\
AIRIA\\
China\\
{\tt\small wangfeng@airia.cn}
\and
Qinghao Hu\\
AIRIA\\
China\\
{\tt\small huqinghao@airia.cn}
\and
Cong Leng\\
AIRIA\\
China\\
{\tt\small lengcong@airia.cn}
}

\author{Xianyang Li\quad Feng Wang \quad Qinghao Hu \quad Cong Leng\\
		Nanjing Artificial Intelligence Chip Research, Institute of Automation, Chinese Academy of Sciences(AIRIA)\\
       China\\
		{\tt\footnotesize qq1052308104@gmail.com \quad \{wangfeng, huqinghao, lengcong\}airia.cn}}

\maketitle
\ificcvfinal\thispagestyle{empty}\fi

\begin{abstract}
 With the development of convolutional neural network, significant progress has been made in computer vision tasks. However, the commonly used loss function softmax loss and highly efficient network architectures for common visual tasks are not as effective for face recognition. In this paper, we propose a novel loss function named Li-ArcFace based on ArcFace. Li-ArcFace takes the value of the angle through a linear function  as the target logit rather than through cosine function, which has better convergence and performance on low dimensional embedding feature learning for face recognition. In terms of network architecture, we improved the the perfomance of MobileFaceNet by increasing the network depth, width and adding attention module. Besides, we  found  some useful training tricks for face recognition. Under all the above effects, we won the second place in the deepglint-light challenge of LFR2019~\cite{27}.
\end{abstract}

\section{Introduction}

\begin{figure}[t]
\begin{center}
\includegraphics[height=5cm, width=5.33cm]{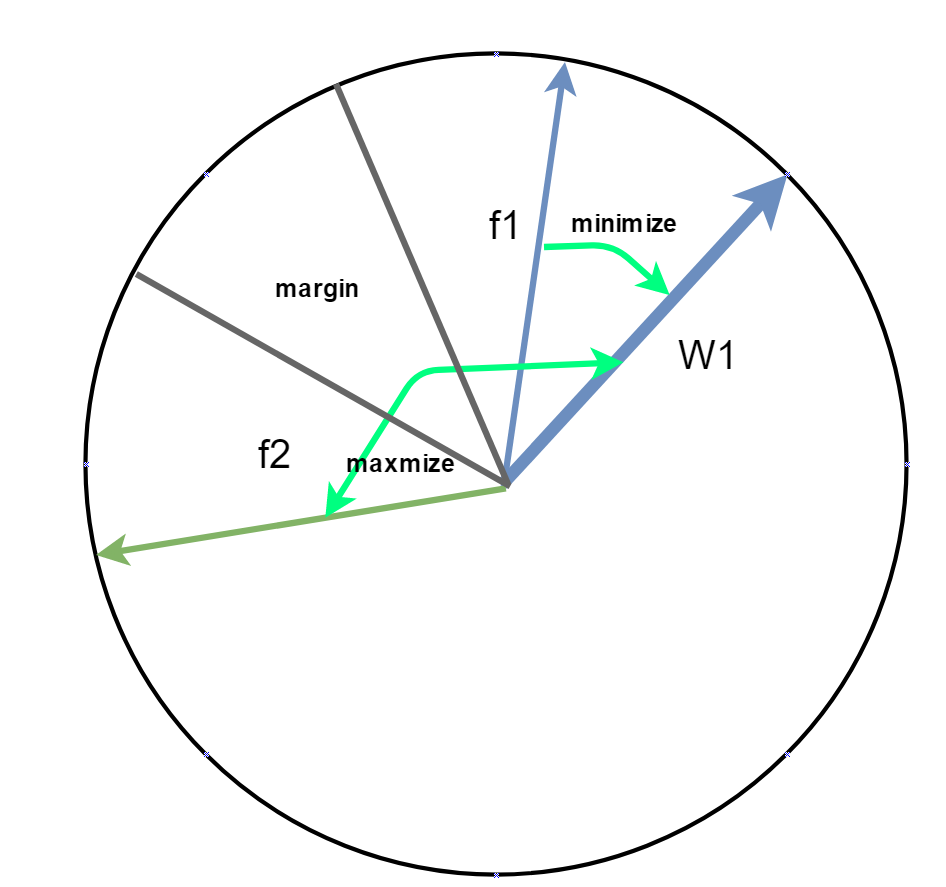}
\end{center}
   \caption{Schematic diagram of discriminative embedding features. W1 refers to the center of the class 1, f1 refers to the embedding feature of class 1, f2 refers to the embedding feature of class 2. During training , the variance of the same person becomes smaller, and the variance of different people becomes larger.}
\label{fig:7}
\end{figure}

The development of deep convolutional networks(DCNN) has made remarkable progress in a series of computer vision tasks.
 However, it is not so effective while  using the common method  for face recognition. Softmax loss, which is commonly used in classification, cant't maximize inter-class variance and minimize intra-class variance of  embedding feature vectors. In order to obtain highly discriminative embedding features(See Figure~\ref{fig:7}), a series of novel loss functions have been proposed in recent years, such as A-Softmax~\cite{3}, CosFace/AM-Softmax~\cite{4, 5}, ArcFace~\cite{6}. Among them, ArcFace achieved state-of-the-art performance by adding additive margin between classes in the angle space. But, when ArcFace is used to train some  efficient networks with small(128) embedding size, it's hard to train from scratch(for  example, the embedding feature size of MobileFaceNet~\cite{7} is 128, it can't converge when trained from scratch with ArcFace). As a result, to guarantee the convergence of training, pre-training is always required with softmax loss. To overcome this problem, we carefully designed a novel loss function named Li-ArcFace based on ArcFace, which performs better in convergence.

 Face recognition technology is now widely used on mobile devices, which requires that the computational cost of the model should not be too large. In recent years, some highly efficient neural network architectures have been proposed, such as MobileNetV1~\cite{17}, ShuffleNet~\cite{18}, and MobileNetV2~\cite{19}, but they're all designed for common visual recognition tasks instead of face recognition. Their performance is really general  when used for face recognition directly. MobileFaceNet is designed for face recognition based on MobileNetV2, achieved remarkable accuracy on LFW~\cite{8}, AgeDB~\cite{9}. And it is even comparable to state-of- the-art big DCNN model on MegaFace~\cite{10} Challenge 1 under the much smaller computational resources. In this paper, with limited amount of computation, we carefully designed a higher performance network architecture based on MobileFaceNet.

\section{Related Work}

\textbf{Loss function.} Softmax loss is the most commonly used loss function  in classification, which is presented as:
\begin{equation}
L_{1}=-\frac{1}{N} \sum_{i=1}^{N} \log \frac{e^{W_{y_{i}}^{T} x_{i}+b_{y_{i}}}}{\sum_{j=1}^{n} e^{W_{j}^{T} x_{i}+b_{j}}}
\end{equation}
where  $x_{i}$ denotes the embedding feature of the $i$-th sample belonging to the $y_{i}$-th class, and the dimension of the embedding feature(hereinafter referred to as embedding size) is set as $d$. $W_{j} \in \mathbb{R}^{d}$  denotes the $j$-th column of the weight $W \in \mathbb{R}^{d \times n}$  and $b_{j} \in \mathbb{R}$ is the bias term. The batch size is $N$ and the class number of  training data is  $n$. However, the traditional softmax loss lacks the power to supervise the embedding feature to minimize inter-class similarity and maximize intra-class similarity.
In SphereFace~\cite{3} and NormFace~\cite{12},  the bias term is being removed at first and then fixing the $\left\|W_{j}\right\|=1$, $\left\|x_{i}\right\|=s$ by $l_{2}$ normalisation, such that the logit is:
 \begin{equation}
W_{j}^{T} x_{i}=\left\|W_{j}\right\| \times\left\|x_{i}\right\| \times \cos \theta_{j}=s \times \cos \theta_{j}
\end{equation}
Where $\theta_{j}$ denotes the angle between $x_{i}$  and $W_{j}$.  Thus the logit is only depend on the cosine of the angle.   The modified loss function can be formulated as :
\begin{equation}
L_{2}=-\frac{1}{N} \sum_{i=1}^{N} \log \frac{e^{s \cos \theta_{y_{i}}}}{\sum_{j=1}^{n} e^{s \cos \theta_{j}}}
\end{equation}
Although $L_{2}$ guarantees a high similarity of features of the same person, it can't separate different classes very well.  In this paper, we use N-Softmax denotes $L_{2}$.
In ArcFace~\cite{6}, the authors added an additive angular margin $m$ within $\cos \theta_{y_{i}}$, which can enhance the intra-class compactness and inter-class discrepancy, the ArcFace is formulated as :
\begin{equation}
L_{3}=-\frac{1}{N} \sum_{i=1}^{N} \log \frac{e^{s\cos \left(\theta_{y_{i}}+m\right)}}{e^{s\cos \left(\theta_{y_{i}}+m\right)}+\sum_{j=1, j \neq y_{i}}^{n} e^{s \cos \theta_{j}}}
\end{equation}
When using ArcFace to train models with 512-dimensional embedding feature, it has well convergence and state-of-the-art performance. However, it will be  difficult to converge while training some highly efficient networks with 128-dimensional embedding feature from scratch.

\textbf{network  architectures.} Face recognition is being  used  more and more on mobile devices. So it's really important to optimize the trade-off between accuracy and computational cost when designing deep neural network architecture. In recent years, some highly efficient neural network architectures have been proposed for common visual  tasks. MobileNetV1~\cite{17} used depthwise separable convolution instead of traditional convolution to reduce computational cost and improve network efficiency. MobileNetV2~\cite{19}  introduced  inverted residuals and linear bottlenecks to further improve network efficiency. However, these lightweight network architectures  are not so accurate when using these unchanged for face recognition. The author of MobileFaceNet~\cite{7} found the weakness of common mobile networks for Face recognition, and solved it by replacing global average pooling with global depthwise convolution(GDC). And the network  architecture of MobileFaceNet is specifically designed for face recognition with smaller  expansion factors in bottlenecks and more output channels at the beginning of the network architecture.
\begin{figure}[t]
\begin{center}
\includegraphics[height=6.75cm, width=9.0cm]{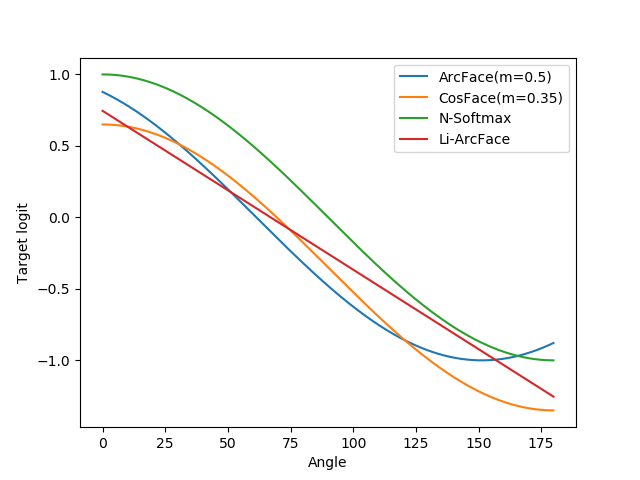}
\end{center}
   \caption{Target logit curves.}
\label{fig:1}
\end{figure}

\begin{figure}[t]
\begin{center}
\includegraphics[height=3cm, width=9.0cm]{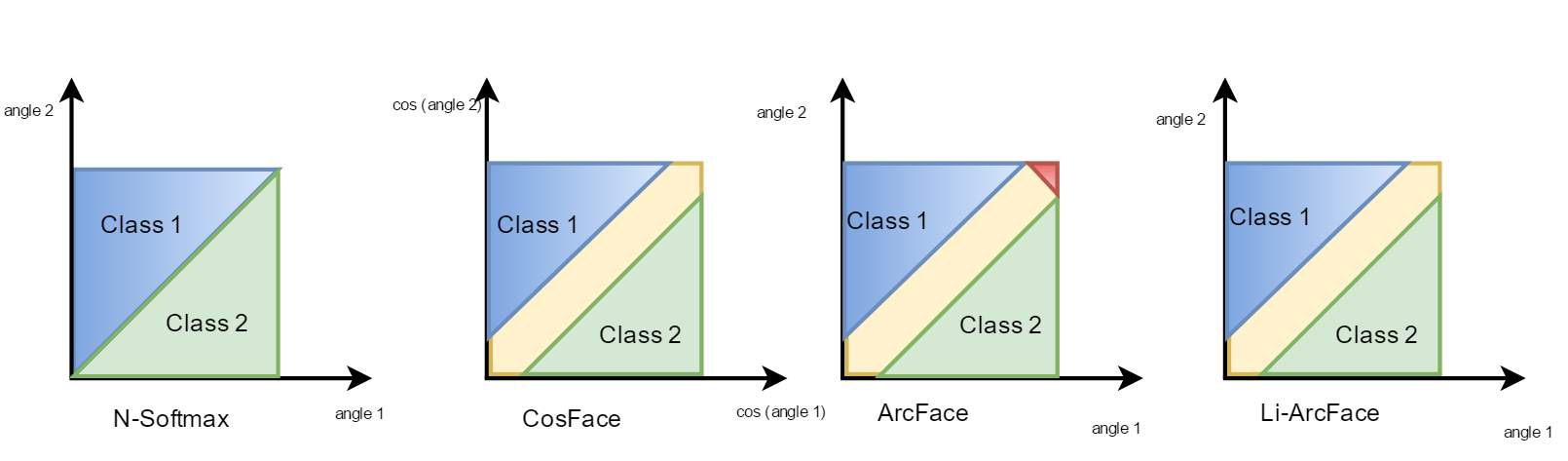}
\end{center}
   \caption{Decision margins of different loss functions under binary classification case. The yellow areas are the decision margins, the red areas are the overlap area of class1 and class2.}
\label{fig:2}
\end{figure}
\section{Proposed Approach}
\subsection{Li-ArcFace}
In ArcFace, the authors added an angular margin $m$ within $\cos\theta_{y_{i}}$, which takes $s\times\cos (\theta_{y_{i}}+m)$ as the target logit. The loss  function proposed by us   takes the angle after a linear function  as the logit rather than cosine function . In the same way, we remove the bias term  and then fix the $\left\|W_{j}\right\|=1$, $\left\|x_{i}\right\|=s$ by $l_{2}$ normalisation,  the $\theta_{j}$ denotes the angle between  $x_{i}$  and $W_{j}$ , thus $\theta_{j}=\arccos \left(W_{j}^{T} x_{i}/s\right)\in[0, \pi]$. At first, we constructed a linear function $f(x)=(\pi-2 x) / \pi$, and we have $f(\theta_{j}) \in[-1, 1] $. Then we add an additive angular margin $m$ in the target logit. In the end we have $s\times(\pi-2 (\theta_{y_{i}}+m)) / \pi$  as the target logit. We call this novel loss function Li-ArcFace, the prefix Li refers to the linear function. The whole Li-ArcFace can be formulated as follows
\begin{equation}
L_{4}=-\frac{1}{N} \sum_{i=1}^{N} \log \frac{e^{s\left(\pi-2\left(\theta_{y_{i}}+m\right)\right) / \pi}}{e^{s\left(\pi-2\left(\theta_{y_{i}}+m\right) \right)/ \pi}+\sum_{j=1, j \neq y_{i}}^{n} e^{s\left(\pi-2\theta_{j}\right) / \pi}}
\end{equation}

There are two main advantages of using this linear function to replace cosine function. Firstly, it is  monotonic decreasing when the angle is between $0$ and $\pi+m$, which will have better convergence, especially for model with small embedding size. For example, when training MobileFaceNet with ArcFace, it will lead to  divergence(NaN). Therefore, softmax loss must be used for pre-training before convergence. The proposed loss does not require these two-stage training. Secondly, the penalty of the proposed loss function  increases linearly as the angle between embedding feature $x_{i}$ and center $W_{y_{i}}$ increasing,  so that the target logit decreases linearly which is more intuitive(See Figure~\ref{fig:1}). It will not decline rapidly in some angles, but slowly in others (corresponding to the gradient value of the target logit curve), which makes the proposed loss function have better performance in the end. In terms of geometric decision margins, ArcFace will have a part of overlap area when the $x_{i}$ deviates too much from the center $W_{y_{i}}$, because of its non-monotonicity of the target logit curve. This area can be distinguished as class 1 and class 2 with ArcFace. The proposed loss function would't have this overlap area(See Figure ~\ref{fig:2}).

\subsection{Network  Architectures}
\begin{table}
\begin{center}
\begin{tabular}{|c|c|c|c|c|c|}
\hline
Input &	Operator&	t	&c&	n	&s\\
\hline\hline
112x112x3	&conv 3x3&	-	&64	&1&	2\\

56x56x64	&depthwise conv3x3	&-	&64	&1	&1\\
56x56x64	&bottleneck	&2	&64	&1	&2\\
28x28x64	&bottleneck	&2	&64	&9	&1\\
28x28x64	&bottleneck	&4	&128 &1	&2\\
14x14x128	&bottleneck	&2	&128	&16	&1\\
14x14x128	&bottleneck &8	&256	&1	&2\\
7x7x256	&bottleneck	&2	&256	&6	&1\\
7x7x256	&conv1x1&	 -	&1024	&1	&1\\
7x7x1024	&linear GDConv7x7&	-	&1024	&1	&1\\
1x1x1024	&linear conv1x1&	-	&512	&1	&1\\

\hline
\end{tabular}
\end{center}
\caption{ The proposed network architecture, n refers to the number of repetitions, c refers to output channels, t refers to The expansion factor.}
\label{tab1}
\end{table}
In this section, we introduce our proposed network architecture. Our network architecture is based on a deeper MobileFaceNet(y2)~\cite{6}, so the residual bottlenecks proposed in MobileNetV2 are used as our main building blocks. Table~\ref{tab1} shows the details of our network architecture. We follow  the  MobileFaceNet,  expansion factors for bottlenecks in our architecture are much smaller than those in MobileNetV2 and  using PReLU as the non-linearity rather than Relu. Nevertheless, we  noticed the importance of network width in  face recognition. MobileFaceNet is significantly wider than MobileNetV2 at the beginning of the network. But MobileFaceNet does not double the output channels of the network when sampling  at last stage. We did that  within  limited amount of computation. At the same time, we carefully adjusted the depth of the network, and  we introduced the attention module CBAM \cite{20} into every bottleneck in the network. But we changed the last activation function in the attention module from sigmoid to 1+tanh. The range of 1+tanh is 0 to 2. When a channel or spatial position needs to be enhanced, it is multiplied by a weight between one and two; when it needs to be weakened, it is multiplied by a weight between zero and one. Using 1+tanh instead of sigmoid is more intuitive and makes training converge faster.

\subsection{Training Tricks for Face Recognition}
During the competition of LFR, we found some useful training tricks for face recognition. Firstly, using a variety of loss functions to fine-tune the model will make the features more robust and improve the accuracy to some extent. In the competition, we used Li-ArcFace, ArcFace, combined loss to fine-tune our model. Secondly, in 512-dimensional embedding feature space, it is difficult  for the lightweight  model  to learn  the distribution of the features. It is an effective method to use some large models to guide the feature distribution of lightweight  models~\cite{14, 15}.
\section{Experiments}
\subsection{Ablation Experiment of Li-ArcFace}
In this section, we mainly introduce some comparative experiments of different loss functions.

\textbf{Implementation Details.} We use the MobileFaceNet as the network architecture, in which the embedding size is set as 128. And batch size is set as 256 x 4. Training models on four NVIDIA TITIAN XP GPUs. We use SGD with momentum 0.9 to optimize models. The scale of the feature is set as 64. The training data set is CASIA- Webface, which contains 10K Identity  0.5M pictures. The learning rate starts at 0.1, divides by 10 at 18k, 26k and 29K iterations, and finally stops training at 30K iterations. At last, we compare the performance on Labelled Faces in the Wild (LFW)~\cite{8}, Celebrities in Frontal Profile (CFP)~\cite{13} and Age Database (AgeDB)~\cite{9}. In this paper, ArcFace (m=0.5, NS) refers to the margin of ArcFace is set as 0.5, and pre-training via N-Softmax before using ArcFace.

\begin{table}
\begin{center}
\begin{tabular}{|l|ccc|}
\hline
Weight decay & LFW&CFP-FP & AgeDB \\
\hline\hline
5e-4 &99.17&	\textbf{94.09}& 92.95 \\
5e-4 wdml10 &\textbf{99.18}& 93.86& \textbf{93.18} \\
4e-5 wdml10 &99.05& 92.07& 92.00 \\
\hline
\end{tabular}
\label{tab:1}
\end{center}
\caption{Verification performance(\%) of different weight decay values. Wdml10 represents the  weight decay multiplier (wd\_mult) parameter of the last convolution layers is 10.  }
\label{tab2}
\end{table}
\textbf{Weight decay.} Before we start the comparison of different loss functions, we first do the numerical experiment of weight decay, and finally determine that the weight decay is set to 5e-4,  except the weight decay parameter of the last layer to the embedding  layer being 5e-3. According to the experimental results(See Table ~\ref{tab2} ), the weight decay of the last layer is more demanding.

\textbf{Effect of Hyper-parameter m. }In Table ~\ref{tab3}, we firstly explored the optimal setting for $m$ of Li-ArcFace, and then we found it was between 0.4 and 0.45.  We tried training model from scratch with ArcFace, but diverged after about 1600 iterations. Therefore, all the experiments with ArcFace  were pre-trained by N-Softmax. Since the embedding size is smaller  than 512, we also explored the optimal setting for $m$ of ArcFace, and we found it was between 0.45 and 0.5.
\begin{table}
\begin{center}
\begin{tabular}{|l|ccc|}
\hline
Loss & LFW&CFP-FP & AgeDB \\
\hline\hline
N-Softmax	&97.87	&90.46	&86.55\\
\hline
Li-ArcFace(m=0.35)	&99.20	&94.20	&92.87\\
\textbf{Li-ArcFace(m=0.4)}	&\textbf{99.27}	&94.11	&\textbf{93.25}\\
Li-ArcFace(m=0.45)	&99.23	&\textbf{94.20}	&93.15\\
Li-ArcFace(m=0.5)	&99.20	&94.10&	92.93\\
\hline
Li-ArcFace(m=0.4, NS)&99.23	&\textbf{94.51}	&92.90\\
Li-ArcFace(m=0.45, NS)&99.27	&94.39	&93.22\\
\hline
ArcFace(m=0.4, NS)&99.27	&93.93	&92.72\\
ArcFace(m=0.45, NS)&99.27	&94.00	&92.98\\
ArcFace(m=0.5, NS)&99.18	&93.86	&93.18\\
\hline
CosFace/AM-Softmax&99.20	&93.39	&92.55\\

\hline
\end{tabular}
\end{center}
\caption{Verification performance (\%) of different loss functions. ArcFace (m=0.5, NS) refers to the margin of ArcFace is set as 0.5, and pre-training via N-Softmax before using ArcFace. }
\label{tab3}
\end{table}

\begin{figure}

\begin{minipage}[t]{0.5\linewidth}
\centering
\includegraphics[width=4.5cm]{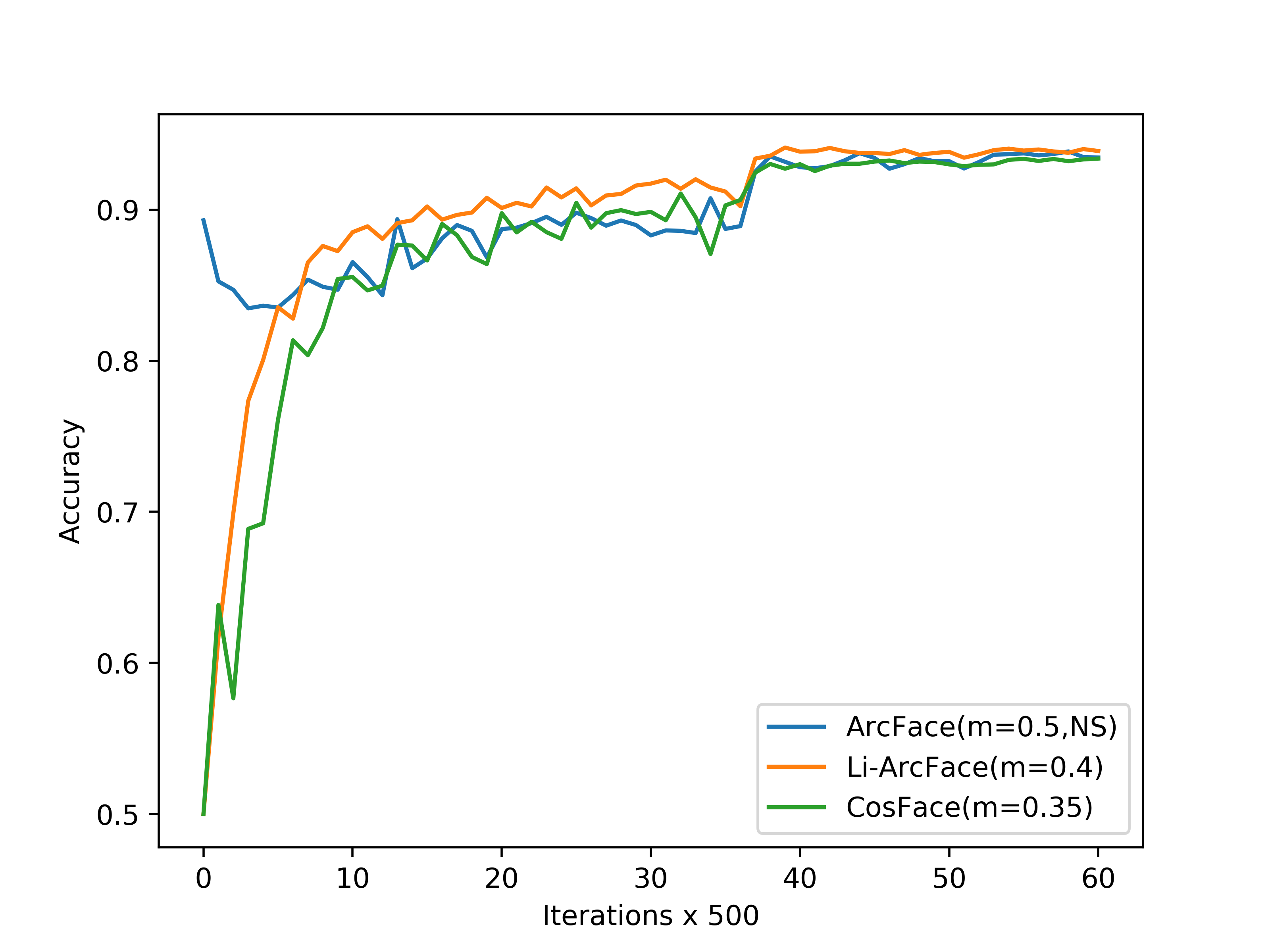}

\end{minipage}%
\begin{minipage}[t]{0.5\linewidth}
\centering
\includegraphics[width=4.5cm]{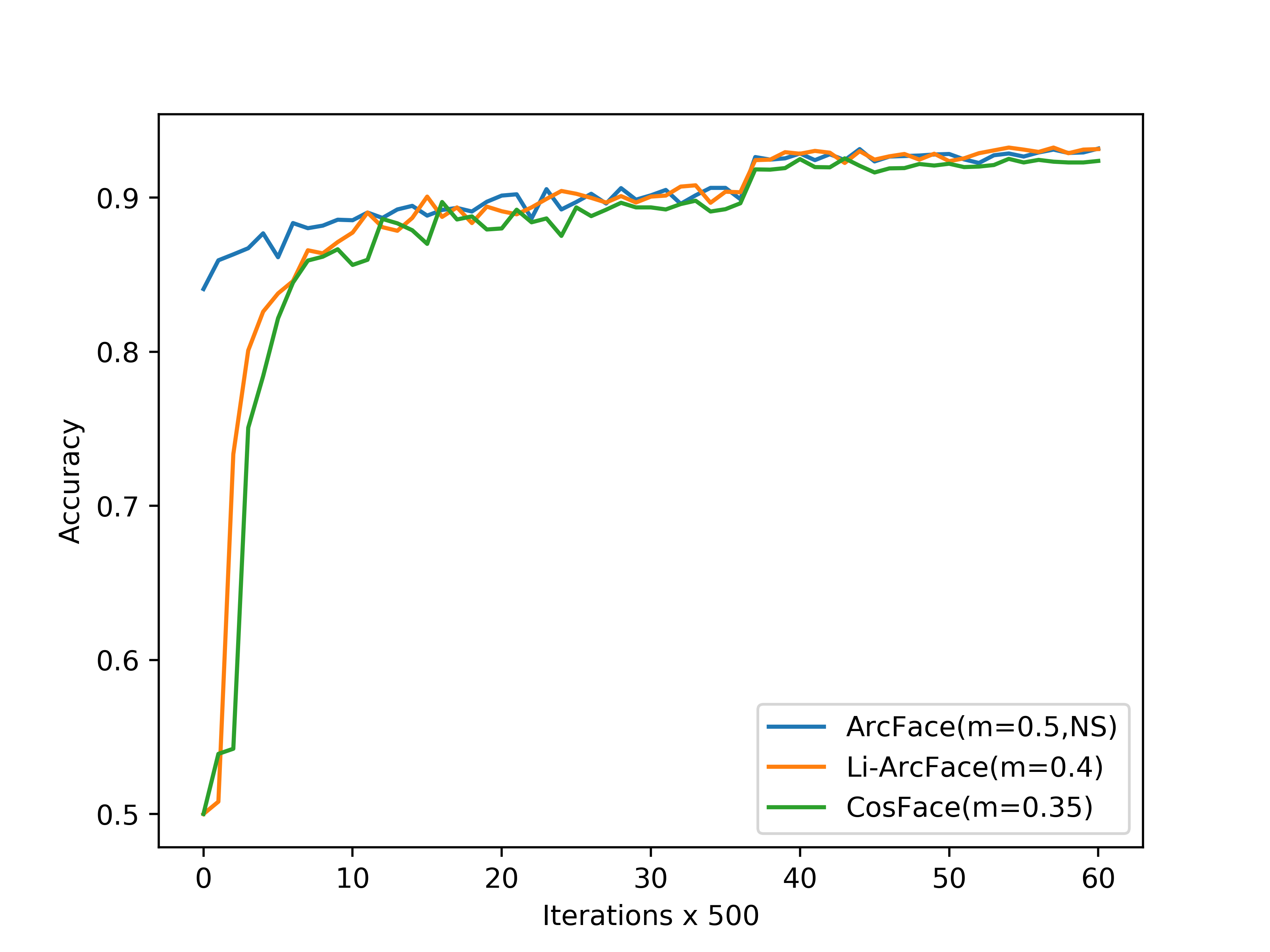}

\end{minipage}
\caption{ The accuracy of the verification sets during training. CFP-FP is on the left, AgeDB is on the right.}
\label{fig:3}
\end{figure}

\textbf{Comparison with state-of-the-art loss functions.} In Table ~\ref{tab3}, the difference between Li-ArcFace, ArcFace and Cosface is tiny on LFW, but all of them are obviously better than N-Softmax;  On CFP-FP and AgeDB,  Li-ArcFace is slightly better than ArcFace and CosFace. We have drawn the accuracy On CFP-FP and AgeDB during the training, which makes the contrast more obvious(See Figure~\ref{fig:3}). We also  compared  Li-ArcFace and ArcFace in the same situation pre-trained by N-Softmax.  On CFP-FP, Li-ArcFace achieves the highest accuracy, and Li-ArcFace is still slightly better than ArcFace on AgeDB. In general, there is little difference in accuracy on the verification sets, but Li-ArcFace has better convergence and does not need pre-training stage when training the model with small embedding size.
\subsection{Evaluation Results of  Network Architecture and Training Tricks}
We name our model that contains our network architecture and training tricks as AirFace. Under the same training data set MS1M-retina~\cite{22}    and model constraints, ( MS1M-retina is cleaned from ~\cite{24} by ~\cite{6}, and ~\cite{26} is the face detector and alignment tool  used to pre-process the data)    the accuracy of AirFace reached 88.415\%@FPR=1e-8 in deepglint-light challenge of LFR19~\cite{27,16}. Meanwhile, we verified the performance of AirFace in MegaFace Challenge 1 compared with the previous state-of-the-art models.  In Table~\ref{tab4}, AirFace has reached incredible efficiency and performance.
\begin{table}
\begin{center}
\begin{tabular}{|l|ccc|}
\hline
Methods & Id(\%)&Ver(\%) & Flops \\
\hline\hline

FaceNet~\cite{21}	    &    70.49	&86.47	& -	\\	
CosFace~\cite{4}	   &     82.72 &	96.65& - \\			
R100,MS1MV2,ArcFace~\cite{6}	&81.03&	96.98	&    27G	\\
R100,MS1MV2,CosFace~\cite{6}	&80.56&	96.56	&    27G	\\
R100,MS1MV2,ArcFace,R~\cite{6}&	\textbf{98.35}&	\textbf{98.48}&	27G\\
R100,MS1MV2,CosFace,R~\cite{6}&	97.91&	97.91&	27G\\
MobileFaceNet~\cite{7}      &-&              90.16&440M\\
MobileFaceNet,R~\cite{7} &-&92.59&440M\\
\hline
AirFace,MS1M-retina    &   80.80 &   96.52   &    1G\\
AirFace,MS1M-retina,R	  &   \textbf{98.04}& 	\textbf{97.93}	&     \textbf{1G}\\

\hline
\end{tabular}
\end{center}
\caption{Identification and verification evaluation on MegaFace Challenge1. Id refers to the rank 1 face identification accuracy with 1M
distractors, and Ver refers to the face verification TAR under $1e-6$ FAR. R refers to data refinement on both probe set and 1M distractors. The list of data cleansing and the code for calculating flops are from InsightFace~\cite{23} }
\label{tab4}
\end{table}

\section{Conclusions}

In this paper, first of all, we propose a novel additive margin loss function for deep face recognition based on ArcFace. The proposed loss function solves the problem that ArcFace does not converge in training model with small embedding feature size. And it achieves the state-of-the-art results on several face verification datasets. Second, we have carefully designed an efficient network architecture and explored some useful training tricks for face recognition, which makes our model AirFace extremely efficient at  both deepglint-light challenge and MegaFace Challenge 1.

{\small
\bibliographystyle{ieee}
\bibliography{egbib}
}

\end{document}